\documentclass[Afour,times,sageapa]{sagej}

\usepackage{amsmath,amssymb}
\usepackage{graphicx}
\usepackage{booktabs}
\usepackage{tikz}
\usetikzlibrary{arrows.meta,positioning,fit,backgrounds}
\usepackage{url}
\usepackage[hidelinks]{hyperref}
\usepackage{cleveref}

\usepackage{xpatch}

\newcommand{\preprintdate}{July 2026}

\makeatletter

\xpatchcmd{\@maketitle}
  {\hbox{}\scriptsize\journalname\\
   \hbox{}\volumenumber(\issuenumber):\startpage--\endpage\\
   \hbox{}\copyright The Author(s) \volumeyear\\
   \hbox{}Reprints and permission:\\
   \hbox{}sagepub.co.uk/journalsPermissions.nav\\
   \hbox{}DOI: 10.1177/ToBeAssigned\\
   \hbox{}www.sagepub.com/\\[2.3pt]
   \hbox{}{\fboxsep 1.5pt\framebox[14mm]{{\normalsize SAGE}}}}
  {\hbox{}\scriptsize\textbf{Preprint}\\
   \hbox{}Submitted to \textit{Adaptive Behavior}\\
   \hbox{}Author manuscript\\
   \hbox{}\preprintdate\\[2.3pt]
   \hbox{}{\fboxsep 1.5pt\framebox[18mm]{{\normalsize PREPRINT}}}}
  {}
  {\PackageWarning{preprint}{Could not replace first-page metadata}}

\def\ps@sagepage{%
  \let\@mkboth\@gobbletwo
  \def\@evenhead{\parbox{\textwidth}{%
    \normalsize\sagesf\thepage\hfill\itshape Preprint\\[-6pt]
    \noindent\rule{\textwidth}{0.25pt}}}%
  \def\@oddhead{\parbox{\textwidth}{%
    \normalsize\sagesf{\itshape\leftmark}\hfill\thepage\\[-6pt]
    \noindent\rule{\textwidth}{0.25pt}}}%
  \def\@evenfoot{}%
  \let\@oddfoot\@evenfoot
}

\def\ps@title{%
  \def\@oddhead{\parbox{\textwidth}{%
    \mbox{}\\[-1pt]
    \noindent\rule{\textwidth}{0.5pt}}}%
  \let\@evenhead\@oddhead
  \def\@oddfoot{}%
  \let\@evenfoot\@oddfoot
}

\makeatother

\begin{document}

\runninghead{Pae}
\title{Same World, Differently Given: History-Dependent Perceptual Reorganization in Artificial Agents}

\author{Hongju Pae\affilnum{1}}
\affiliation{
\affilnum{1}Active Inference Institute, Crescent City, CA, USA
}
\corrauth{
  Hongju Pae, Active Inference Institute, Crescent City, CA, USA
  }
\email{hjpae@activeinference.institute}

\begin{abstract}
What kind of internal organization would allow an artificial agent not only to adapt its behavior, but to sustain a history-sensitive perspective on its world? I present a minimal architecture in which a slow perspective latent $g$ feeds back into perception and is itself updated through perceptual processing. This allows identical observations to be encoded differently depending on the agent's accumulated stance.
The model is evaluated in a minimal gridworld with a fixed spatial scaffold and sensory perturbations. Across analyses, three results emerge. First, the perspective latent reorganizes perceptual encoding: identical observations are represented differently depending on prior experience, and this reorganization of salience gating replicates across runs, with five of 16 gating dimensions changing direction consistently across 30 independent runs after correction for multiple comparisons. Second, only adaptive self-modulation yields the characteristic growth-then-stabilization dynamic of the perspective latent, unlike rigid or always-open update regimes. Third, perturbation history is followed by reduced adaptive plasticity after nominal conditions are restored, showing a directionally consistent trend across seeds. Gross behavior remains stable throughout the analysis, suggesting that the dominant reorganization is perceptual rather than behavioral. 
Together, these findings identify a minimal mechanism for history-dependent perspectival organization in artificial agents.
\end{abstract}


\maketitle

\section{Introduction}

\par The same nominal stimulus need not be encountered as the same world. A cue, a pattern, or an event may be formally identical at the sensory level and yet be taken up very differently depending on the history through which it is encountered: as familiar or anomalous, negligible or significant, safe or threatening. What changes in such cases is not the external input alone, but the mode in which it is disclosed to the perceiving subject. This paper investigates what kind of \textit{internal organization} would allow an artificial agent to inhabit a stable and history-sensitive perspective on its world. To this end, it introduces a minimal agent architecture in which accumulated experience feeds back into perception itself.

\par The question has deep roots in phenomenology. A subjective viewpoint is not exhausted by accurate world-estimates; instead, lived experience is structured by a \textit{perspective} under which the world appears as inviting, threatening, neutral, or significant in other salient qualitative ways \cite{merleau1945,thompson2007,pae2026perspective}. In Husserlian terms, intentionality is characterized not only by intentional matter (what the mind is about), but also by the \textit{intentional quality}, or the underlying stance-like mode in which it is given \cite{husserl1913}. Contemporary phenomenology has continued to treat such perspectival organization as a basic condition for situated cognition \cite{gallagher2020,gallagher2023}. On this view, perspective is more than just a memory of prior states; it is part of the condition under which a situation appears as threatening, negligible, familiar, or significant in the first place.

\par Because ``perspective'' is used in several distinct senses across cognitive science, it is worth fixing terminology at the outset. The present work is not concerned with social perspective-taking, i.e. the capacity to represent another agent's viewpoint as studied in Theory-of-Mind research. Instead, ``perspective'' is used here in a first-person, structural sense: the agent's own accumulated stance, which conditions how its world is given to it. Correspondingly, \textit{perspectival perception} refers throughout to the property that the same nominal observation can be encoded differently depending on this accumulated stance. Both notions are made operationally precise below: perspective is identified with a slow latent variable $g$, and perceptual reorganization with measurable, history-dependent changes in how a frozen encoder's output is modulated.

\par This broader phenomenological view has increasingly influenced computational work at the intersection of enactivism-based embodied cognition and artificial life. Recent enactive and embodied AI research has argued that intelligent systems may require forms of organization that are irreducible to static representations, pre-specified objectives, or centralized control \cite{dreyfus2007,froese2009}. On this view, adaptive intelligence depends on ongoing agent-environment coupling, history-sensitive self-organization, and the maintenance of internal coherence across perturbation and change \cite{varela1991,dipaolo2005,ziemke2009,linson2018}. Research on artificial life and adaptive behavior has long served as a productive meeting ground where such questions about minimal cognition can be investigated through constructive simulation rather than post-hoc interpretation \cite{froese2010,beer2003,kirchhoff2017}.

\par In prior work \cite{pae2026}, I introduced a slow global latent variable $g$ that operates on a timescale complementary to the action policy. In that work, $g$ exhibited temporal hysteresis extending beyond the timescale of policy adaptation: in a reward-free regime-switching environment, $g$ lagged behind shifts in environmental conditions by tens of episodes, remaining dissociated from a policy that adapted on a much faster timescale. These results suggested that such a slow latent may function less as an instantaneous action-related variable and more as a persistent stance regarding \textit{what ``kind'' of world} the agent continues to assume it inhabits.

\par That work left open, however, whether such a latent can do more than passively accumulate perceptual history. The present study therefore asks: \textbf{can a slow latent actively reorganize perception itself, in a history-dependent manner?} If perspective is genuinely a condition of givenness rather than a mere record of the past, then prior experience should not only be retrievable; it should change how subsequent observations are encoded in the first place.

\par The architecture presented here is designed to operationalize exactly this possibility. Rather than treating perception as a fixed preprocessing stage and the slow latent as a passive summary of prior observations, the model allows prior perspective to feed back into perceptual encoding itself. Concretely, I extend the architecture introduced in prior work \cite{pae2026} in two ways. First, \textit{salience gating}: the prior perspective latent $g_{t-1}$ modulates observation encoding through feature-wise linear modulation (FiLM), following Perez et al. \citeyear{perez2018film}, so that the same raw input can be encoded differently under different accumulated perspectives. Second, \textit{self-modulating plasticity}: the latent helps determine its own openness to revision through an adaptive update rate, inspired by neuromodulated plasticity in biological neural networks \cite{miconi2019,najarro2020}. Together, these mechanisms close a feedback loop in which perspective shapes perception and gated perception, in turn, conditions how readily perspective itself changes. The mechanisms are evaluated in a minimal gridworld whose spatial scaffold remains fixed while transient perturbations distort sensory input, allowing history effects to be isolated from environmental change.

\par Within this setting, the paper addresses three research questions, each paired with a dedicated experimental analysis:

\paragraph{Perceptual reorganization.} Does the perspective latent $g$ reorganize the encoding of identical observations as a function of history? This is tested with a fixed probe set passed through a frozen encoder, so that any difference in the resulting representations is attributable solely to $g$-driven modulation.

\paragraph{Self-modulating plasticity.} Is adaptive self-modulation of plasticity necessary for stable perspectival organization, or does any fixed update rate suffice? This is tested by ablating the adaptive update law against rigid (slow) and open (fast) fixed-rate controls.

\paragraph{History residue.} Does transient perturbation history leave a measurable residue in the agent's internal organization after nominal conditions are restored? This is tested by comparing self-regulated plasticity across matched three-block schedules that differ only in intervening perturbation experience.

\par The results answer the first two questions in the affirmative and provide consistent directional evidence for the third. The perspective latent reorganizes how the same nominal observations are encoded under controlled intervention and in naturally occurring episode histories, reproducibly across runs; only adaptive self-modulation yields the characteristic growth-then-stabilization dynamic, which neither rigid nor always-open updating reproduces; and intervening perturbation is followed by reduced plasticity between nominally identical blocks, a trend consistent in direction across seeds. 

\par Throughout, the agent's gross behavior remains comparatively stable, indicating that the dominant reorganization is perceptual rather than behavioral. Taken together, these findings identify a minimal mechanism by which experiential history becomes an organizing condition for subsequent experience; a computational analog of the phenomenological claim that perspective belongs to the conditions of givenness.

\subsection{Operational Definitions}
\par This work is motivated in particularly phenomenological vocabulary. To keep such motivation anchored to measurable quantities, the following terms are used in a fixed operational sense throughout the remainder of the paper.

\paragraph{Perspective (state).} The value of the slow latent $g_t$ (Eq.~\ref{eq5}). The phrase \textit{accumulated stance} refers to the same variable, emphasizing that its value at time $t$ reflects the integrated history of prior episodes rather than the current observation alone.

\paragraph{Perspectival perception.} The property that the same nominal observation $x$ is encoded into different perceptual latents $z$ depending on the perspective state $g$ under encounter. This is measured with the fixed-probe assay; since the encoder base MLP is frozen, any difference between $z(g)$ and $z(g')$ for identical probe inputs is attributable solely to $g$-driven FiLM modulation (Eqs.~\ref{eq2}--\ref{eq3}).

\paragraph{Residue.} A measurable difference in the agent's internal variables that persists after nominal environmental conditions have been restored. Residue is operationalized in two ways: (i)~the excess decrease in mean plasticity $\alpha$ between matched no-perturbation blocks, relative to the decrease observed under a no-perturbation baseline schedule of equal duration; and (ii)~the reorganization of the FiLM salience coefficients $\gamma$ between the pre- and post-perturbation no-perturbation blocks. Under this definition, residue becomes a property of the agent's self-organization, but not of the environment, as the environmental scaffold and noise map remains identical in the compared blocks.

\paragraph{Self-modulating plasticity.} The property that the update rate $\alpha_t$ of the perspective state is itself a bounded function of the agent's current state ($z_t$, $p_t$, $g_{t-1}$, $e_t$; Eq.~\ref{eq4}). \textit{Openness to revision} refers to the momentary value of $\alpha_t$.

\section{Agent Architecture}

\par The broad architectural motivation is to prevent perspective from collapsing into the action policy itself. If all internal state is optimized only for immediate action selection, then history-sensitive organization is difficult to distinguish from policy adaptation. The architectural backbone therefore separates a fast action pathway from a slow global latent, allowing the latter to function as a longer-timescale constraint on how the current world is taken up by the agent.

\subsection{Base Architecture}
\par The agent consists of five components inherited from the prior architecture \cite{pae2026}. Throughout, the slow global latent $g$ is referred to as the \textit{perspective latent}. All module dimensions are summarized in Table~\ref{tab:notation}.

\paragraph{Observation encoder.} At each timestep, the agent receives a local observation $x_t \in \mathbb{R}^{8}$, containing the (noisy) values of 8 cells neighboring its current position. A two-hidden-layer MLP (width 64, \texttt{Tanh} activations) maps $x_t$ to a raw perceptual code $z_{\mathrm{raw}} \in \mathbb{R}^{16}$ (Eq.~\ref{eq1}). In the base architecture, the perceptual latent is simply $z_t = z_{\mathrm{raw}}$; the salience-gated version is newly introduced.

\paragraph{Action-trace encoder.} The previous action $a_{t-1}$ (one of five discrete actions) is mapped through a learned embedding to a proprioceptive trace $p_t \in \mathbb{R}^{8}$, giving the agent access to its own recent motor history independently of the observation stream.

\paragraph{Perspective latent.} The perspective latent $g_t \in \mathbb{R}^{12}$ is maintained by a \texttt{GRUCell} that receives the concatenation $z_t \oplus p_t$ as input and $g_{t-1}$ as its hidden state, followed by layer normalization. In the first training stage the update uses a fixed interpolation rate and $g$ is reset at episode boundaries; in the second stage the update is replaced by the self-modulating formulation of Eqs.~\ref{eq4}--\ref{eq5} and $g$ is carried across episode boundaries, so that structure accumulates over the full 150-episode run. Note that this asymmetry is deliberate: stage one establishes a behavioral backbone under within-episode dynamics only, whereas stage two is precisely the regime in which cross-episode history is allowed to matter.

\paragraph{State head and policy.} A state head composes the current percept and stance into an action-relevant state: $s_t = \tanh(\mathrm{LN}(\mathrm{MLP}(z_t, p_t, g_t))) \in \mathbb{R}^{16}$. A discrete action policy $\pi(a_t \mid s_t)$ over the 5 actions (up, down, left, right, stay) is computed from $s_t$ alone, so that the perspective latent influences action selection only through its contribution to $s_t$.

\paragraph{Observation decoder.} An observation decoder predicts the next observation, $\hat{x}_{t+1} = D(g_t, a_t)$, from the current perspective state $g_t$ and the selected action $a_t$. One-step prediction error on this reconstruction is the primary learning signal in both training stages; no extrinsic reward is provided at any point.

\paragraph{Gradient decoupling of the perspective latent.} A central design constraint is that the perspective latent must not be shaped by action-side optimization. The policy therefore operates on a stop-gradient copy of the state, $\pi(a_t \mid \mathrm{sg}(s_t))$, so that actor gradients update the policy network only and cannot propagate into the state head, the encoder, or $g$. Likewise, the smoothness regularizer compares $g_t$ against a detached copy of $g_{t-1}$. As a result, $g$ is shaped exclusively by predictive coherence (one-step reconstruction) and its own temporal regularization, but never by action-selection objectives. This decoupling is what licenses interpreting $g$ as a perceptual-organizational variable rather than a policy-serving hidden state, and it underwrites the comparison with belief states and generic recurrent dynamics developed in the Discussion.

\begin{table*}[t]
\centering
\small
\begin{tabular}{@{}lllll@{}}
\toprule
Var. & Dim. & Role & Timescale & Stage 2 \\
\midrule
$x_t$ & 8 & local observation patch & step & (input only) \\
$z_{\mathrm{raw}}$ & 16 & pre-gating percept & step & frozen \\
$z_t$ & 16 & gated perceptual latent & step & FiLM trainable \\
$p_t$ & 8 & action-trace embedding & step & frozen \\
$g_t$ & 12 & perspective latent & cross-episode & trainable \\
$\alpha_t$ & 1 & plasticity rate $\in[0.03, 0.30]$ & step (state-dep.) & trainable \\
$(\gamma, \beta)$ & $16{+}16$ & FiLM coefficients & fn.\ of $g_{t-1}$ & trainable \\
$s_t$ & 16 & policy state & step & frozen \\
$e_t$ & 6 & error-feedback features & step & (input only) \\
\bottomrule
\end{tabular}
\caption{Notation, dimensionality, functional role, characteristic timescale, and second-stage training status of all model variables. ``Frozen'' components are loaded from first-stage checkpoints and not updated in stage two.}
\label{tab:notation}
\end{table*}

\subsection{Architectural Extensions}
\par While the base architecture established a temporal dissociation between a slow global latent and a fast action policy, it left open whether the latent merely accumulated perceptual history or could actively reorganize perception itself. The present model extends the architecture in precisely this direction. Rather than treating the slow latent as a passive background variable, it allows the latent to shape how observations are encoded and how readily its own state is revised. As illustrated schematically in Fig.~\ref{fig1}, this introduces a feedback loop between the perceptual encoding and accumulated perspective.

\par The main extensions are as follows:
\begin{itemize}
  \item \textbf{Salience gating:} the prior global latent state $g_{t-1}$ modulates encoding of perceptual latent state $z_t$, allowing the same raw observation to be encoded differently under different accumulated perspectives.
  \item \textbf{Self-modulating plasticity:} the gated percept $z_t$, together with the prior perspective $g_{t-1}$, determines the update (plasticity) rate $\alpha_t$ of the perspective latent, so that openness to revision depends partly on the stance already occupied by the agent.
\end{itemize}

\par Functionally, these additions expand the role of the global latent $g$. In the earlier architecture, $g$ primarily served as a slow-evolving history-sensitive variable dissociable from action policy timescale; additionally, in the present architecture, it acts back on the perceptual pathway and retains its own adaptive dynamics. The resulting model is therefore designed to let accumulated perspective reorganize how the same nominal world is perceived by the agent.

\begin{figure*}[t]
\centering
\begin{tikzpicture}[
    font=\small,
    >=Latex,
    box/.style={draw, rounded corners=3pt, thick, align=center,
                minimum height=9mm, inner sep=3pt},
    env/.style={box, draw=gray!50, fill=gray!8, minimum width=16mm},
    enc/.style={box, draw=green!50!black, fill=green!8, minimum width=18mm},
    film/.style={box, fill=green!10, 
                 draw=violet!50!green!50!black, dashed, very thick,
                 minimum width=18mm},
    persp/.style={box, draw=violet!60!black, fill=violet!12, minimum width=20mm},
    alpha/.style={box, draw=violet!40!black, fill=violet!6, minimum width=18mm},
    pol/.style={box, draw=orange!60!black, fill=orange!10, minimum width=18mm},
    arr/.style={-Latex, thick, gray!60!black},
    grarr/.style={-Latex, thick, gray!40},
    varr/.style={-Latex, very thick, violet!90!black},
    plarr/.style={-Latex, very thick, violet!40!black},
    garr/.style={-Latex, very thick, green!60!black},
    darr/.style={-Latex, thick, dashed, gray!40},
    lbl/.style={font=\scriptsize, gray!60!black},
    sub1/.style={draw=gray!30, rounded corners=4pt, inner sep=6pt},
    sub2/.style={draw=green!30, rounded corners=4pt, inner sep=6pt},
    sub3/.style={draw=violet!30, rounded corners=4pt, inner sep=6pt},
    sub4/.style={draw=orange!30, rounded corners=4pt, inner sep=6pt},
]
\node[env] (ptb) at (0, 2.2) {\scriptsize perturbation};
\node[env] (xt)  at (0, 0)   {$x_t$};
\node[enc] (zr) at (2.8, 0)  {$z_{\mathrm{raw}}$};
\node[film] (fm) at (5.2, 0) {\scriptsize FiLM \\\scriptsize salience gate};
\node[enc] (zt) at (7.6, 0)  {$z_t$\\\scriptsize perceptual latent};
\node[pol] (st) at (10.2, 0) {$s_t$\\\scriptsize policy state};
\node[pol] (pi) at (12.6, 0) {$\pi(a_t|s_t)$\\\scriptsize action policy};
\node[pol] (at) at (15.0, 0) {$a_t$\\\scriptsize action};
\node[persp] (g0) at (5.2, -2.9) {$g_{t-1}$\\\scriptsize perspective};
\node[alpha] (al) at (7.6, -2.9) {$\alpha_t$\\\scriptsize plasticity};
\node[persp] (g1) at (10.2, -2.9) {$g_t$\\\scriptsize updated perspective};
\node[sub1, fit=(ptb)(xt), 
      label={[lbl, anchor=south west]north west:\textit{Environment}}] {};
\node[sub2, fit=(zr)(fm)(zt), 
      label={[lbl, anchor=south west]north west:\textit{Perceptual encoding}}] {};
\node[sub3, fit=(g0)(al)(g1), 
      label={[lbl, anchor=south east]south west:\textit{Perspective dynamics}}] {};
\node[sub4, fit=(st)(pi)(at), 
      label={[lbl, anchor=south west]north west:\textit{Action pathway}}] {};
      
\draw[arr]  (xt) -- (zr);
\draw[arr]  (zr) -- (fm);
\draw[arr]  (fm) -- (zt);
\draw[arr]  (zt) -- (st);
\draw[arr]  (st) -- (pi);
\draw[arr]  (pi) -- (at);
\draw[grarr] ([xshift=-1.0mm]ptb.south) -- 
    node[left, lbl] {\textbf{perturbs}} 
    ([xshift=-1.0mm]xt.north);


\draw[varr] (g0) -- (fm) 
    node[midway, left, font=\scriptsize, violet!90!black, align=center] 
    {\textbf{perspective}\\\textbf{gates perception}};
    
\draw[garr] (zt.south) -- ++(0, -0.6) -| (al.north);
\node[font=\scriptsize, green!60!black, align=center] 
    at (6.55, -1.44) {\textbf{gated perception}\\\textbf{controls plasticity}};
    
\draw[arr, gray!40] ([xshift=-3.0mm]zt.south east) -- ++(0, -0.8) -| ([xshift=-2.5mm]g1.north);

\draw[plarr] ([xshift=2mm]g1.north) -- ([xshift=2mm]st.south)
    node[pos=0.4, right, lbl, violet!60!black, align=center]
    {\textbf{perspective biases}\\\textbf{action selection}};

\draw[arr] (g0) -- (al);
\draw[arr] (al) -- (g1);

\draw[arr, gray!40] (g0.south) -- ++(0, -0.4) -| (g1.south)
    node[pos=0.5, below, lbl] {EMA carry};

\draw[darr] (at.south) -- ++(0, -0.5) -| ([xshift=5.0mm]st.south)
    node[pos=0, below, lbl] {$a_{t-1}$ action trace};
    
\draw[darr, gray!25] (at.north) -- ++(0, 0.8) -| ([xshift=4.0mm]xt.north)  
    node[pos=0.3, above, lbl] {embodied effect};
    
\end{tikzpicture}
\caption{
\textbf{Schematic overview of the extended agent architecture.}
Gray, green, purple, and orange denote the environment, perceptual encoding, perspective dynamics, and action pathway. The main additions are the feedback links between perception and perspective: \textit{salience gating} (purple arrow), in which $g_{t-1}$ modulates perceptual encoding through the FiLM gate, and \textit{self-modulating plasticity} (green arrow), in which gated perception influences the plasticity rate $\alpha_t$. Together with the update path $g_{t-1} \rightarrow \alpha_t \rightarrow g_t$, these implement the central loop of the model: perspective shapes perception, and perception regulates perspective. The updated $g_t$ then biases the fast action pathway through $s_t$. Thin black arrows indicate the main forward flow; gray arrows indicate auxiliary or implicit connections. For clarity, the observation decoder, prediction-error features, and explicit proprioceptive pathway are omitted.
}
\label{fig1}
\end{figure*}

\subsection{Implementation Mechanisms}

\paragraph{Salience gating.}
The observation encoder is extended with a FiLM layer conditioned on the perspective latent:
\begin{equation}
    z_{\text{raw}} = \tanh(\text{MLP}(x_t))
    \label{eq1}
\end{equation}
\begin{equation}
    (\gamma, \beta) = \text{Linear}(g_{t-1})
    \label{eq2}
\end{equation}
\begin{equation}
    z_t = (1 + \gamma) \cdot z_{\text{raw}} + \beta
    \label{eq3}
\end{equation}

\par FiLM weights are zero-initialized, so the gating begins as identity and develops structure through learning. As shown in Fig.~\ref{fig1}, this mechanism allows prior perspective to act directly on perceptual encoding before action selection.

\par Unless otherwise noted, all MLP modules use two hidden layers of width 64 with \texttt{Tanh} activations and no dropout; latent dimensions are $z_t \in \mathbb{R}^{16}$, $p_t \in \mathbb{R}^{8}$, $g_t \in \mathbb{R}^{12}$, and $s_t \in \mathbb{R}^{16}$, while the adaptive update network uses a hidden width of 32.

\paragraph{Self-modulating plasticity.}
The global latent update becomes:
\begin{equation}
    \alpha_t = \bar{\alpha} + \tfrac{\Delta\alpha}{2}\tanh\!\bigl(\text{AlphaNet}(z_t,p_t,g_{t-1},e_t)\bigr)
    \label{eq4}
\end{equation}

\par Here, $\text{AlphaNet}$ is a small multilayer perceptron that takes as input the current perceptual latent $z_t$, the proprioceptive trace $p_t$, the prior perspective state $g_{t-1}$, and an error feature vector $e_t \in \mathbb{R}^6$ summarizing recent prediction-error statistics and perturbation signals, and returns a scalar plasticity rate $\alpha_t$. The network's scalar output is mapped affinely into the admissible range by $\bar{\alpha} = (\alpha_{\min}+\alpha_{\max})/2 = 0.165$ and $\Delta\alpha = \alpha_{\max}-\alpha_{\min} = 0.27$, so that $\alpha_t \in [0.03, 0.30]$ by construction. The output bias of $\text{AlphaNet}$ is initialized to $-0.75$, giving an initial plasticity of $\alpha_0 \approx 0.08$. In this way, latent plasticity remains sensitive also to recent prediction-error structure. 

\par Thus, the global latent is now updated by a bounded, state-dependent interpolation between the previous perspective state and a GRU-based candidate update:
\begin{equation}
    g_t = (1 - \alpha_t) \cdot g_{t-1} + \alpha_t \cdot \text{LN}\!\left(\text{GRU}(z_t \oplus p_t,\, g_{t-1})\right)
    \label{eq5}
\end{equation}

\section{Experiment Methods}

\subsection{Simulation Environment}
\par
\begin{figure*}[t]
\centering
\includegraphics[width=1.0\textwidth]{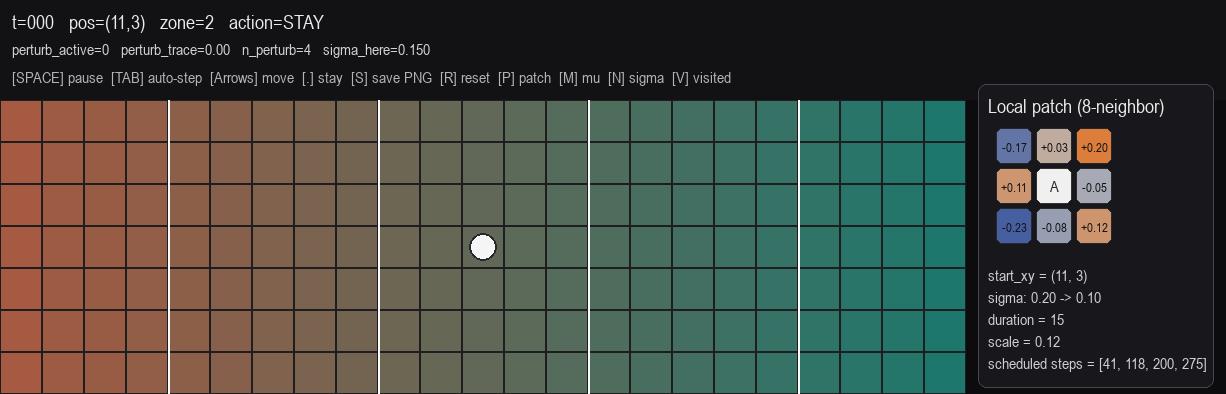} 
\caption{\textbf{Gridworld environment visualization.} This Pygame-based simulator renders a fixed $23 \times 7$ gridworld with a left-to-right observation noise gradient: left-side cells are noisier and right-side cells are more reliable. For analysis, the 23 columns are divided into 5 reporting zones (vertical white lines). Background color indicates noise level, from red (high) to teal (low). The agent (white circle) starts at the grid center. The top overlay shows timestep, position, reporting zone, action, perturbation status and trace, number of scheduled perturbations, and local observation noise $\sigma$. The right panel shows the current 8-neighbor local observation patch $x_t$. This visualization is illustrative only; all simulations are run headless.}
\label{fig2}
\end{figure*}

 As illustrated in Fig.~\ref{fig2}, the agent operates in a fixed $23 \times 7$ gridworld with a left-to-right observation-noise gradient. Each cell $c$ of the grid carries a fixed mean value $\mu(c)$, and every observed cell value is drawn independently at each timestep from a Gaussian distribution, $x \sim \mathcal{N}(\mu(c), \sigma(c)^2)$, where $\sigma(c)$ denotes the standard deviation of this observation noise. $\sigma$ decreases linearly across columns from the left edge to the right edge, so that observations are systematically less reliable on the left side of the grid. The mean-value field $\mu$ is fixed at environment construction (including a small spatial jitter, s.d.\ $0.015$) and never changes thereafter.

\par At each timestep, the agent observes an 8-dimensional local patch over the 8 neighboring cells and selects one of 5 actions (up, down, left, right, stay). No extrinsic reward function is provided; learning is driven solely by next-step prediction error. For analysis, the 23 columns are grouped into 5 reporting zones spanning the noise gradient.

\subsection{Training Protocol}
\par Training proceeds in two stages on the same $23\times7$ grid scaffold but with different noise geometries. In the first stage, a strong left-to-right predictability gradient is imposed ($\sigma = 0.60 \rightarrow 0.03$). In the second stage, the scaffold is retained, but the gradient is weakened ($\sigma = 0.20 \rightarrow 0.10$) and supplemented with transient perturbations.

\paragraph{Stage 1: behavioral backbone.} The first stage trains all components end-to-end, with neither salience gating nor self-modulating plasticity active. The FiLM branch is initialized to the identity transformation, and the perspective update uses a fixed rate. The learning objective combines four terms:

\begin{equation}
\mathcal{L}
= \mathcal{L}_{\mathrm{pred}}
+ w_{\mathrm{sm}}\mathcal{L}_{\mathrm{smooth}}
+ w_{\mathrm{ac}}\mathcal{L}_{\mathrm{actor}}
- w_H H(\pi)
\label{eq:stage1loss}
\end{equation}

\par The primary learning signal is the one-step prediction error of the policy-expected next observation:

\begin{equation}
\mathcal{L}_{\mathrm{pred}}
= \left\lVert \sum_{a \in \mathcal{A}} \operatorname{sg}\!\bigl(\pi(a \mid s_t)\bigr)
D(g_t,a) - x_{t+1} \right\rVert_2^2
\label{eq:prediction-loss}
\end{equation}
Here, the decoder's action-conditional predictions are averaged under a detached copy of the current policy. Consequently, the perspective latent $g_t$ must support prediction marginalized over the agent's action tendencies. 

\par The temporal smoothness regularizer is: 
\begin{equation}
\mathcal{L}_{\mathrm{smooth}}
= \left\lVert g_t-\operatorname{sg}(g_{t-1}) \right\rVert_2^2
\label{eq:smoothness-loss}
\end{equation}
This term penalizes rapid changes in the perspective latent while preventing
gradients from propagating backward through the previous latent state.

\par And the advantage-weighted actor loss is: 
\begin{equation}
\mathcal{L}_{\mathrm{actor}}
= -\hat{A}_t \log \pi\!\left(a_t \mid \operatorname{sg}(s_t)\right)
\label{eq:actor-loss}
\end{equation}
Here, the actor term uses a cost-based advantage. With $c_t$ as the prediction
cost of the realized action, a running baseline $b_t$ and scale $\sigma_t$
are tracked as exponential moving averages ($\beta_{\mathrm{mean}} = 0.98$,
$\beta_{\mathrm{var}} = 0.99$), giving:

\begin{equation}
A_t = b_t - c_t,
\qquad
\hat{A}_t = \mathrm{clip}\!\left(A_t / \sigma_t,\ \pm 3\right)
\label{eq:advantage}
\end{equation}

\par Finally, $H(\pi)$ denotes the entropy of the action distribution that encourages continued exploration during first-stage training. Loss weights are $w_{\mathrm{sm}} = 0.25$ and $w_{\mathrm{ac}} = 0.5$, and the entropy coefficient $w_{H}$ is adapted online toward a target entropy of $0.60 \log |\mathcal{A}|$, bounded to $[0.002, 0.05]$ with initial value $0.01$. The actor term is gated on only after a 12{,}000-step warmup, allowing the predictive backbone and $g$-dynamics to stabilize before actor-side optimization is introduced. Optimization uses \texttt{Adam} (learning rate $3\times10^{-4}$, gradient-norm clipping at $1.0$) for 36{,}000 online steps in 240-step episodes.

\paragraph{Stage 2: perspective-driven feedback.}
In the second stage, first-stage checkpoints are loaded as a frozen behavioral backbone: the encoder base MLP, action-trace encoder, state head, policy, and observation decoder are all fixed (see Table~\ref{tab:notation}). The perspective update is replaced by the self-modulating formulation of Eqs.~\ref{eq4}--\ref{eq5}, and only the FiLM gate and the perspective-latent module (GRU and AlphaNet) remain trainable. The training signal in this stage is the one-step prediction error $\mathcal{L}_{\mathrm{pred}}$ alone; no actor, entropy, or smoothness terms are applied, so all second-stage learning is driven by predictive coherence under the frozen behavioral backbone. 

\par The adaptive rate is bounded to $\alpha_t \in [0.03, 0.30]$. The error-feedback vector $e_t \in \mathbb{R}^{6}$ supplied to AlphaNet is defined in terms of the one-step prediction error $\epsilon_t$ and its short- and long-run exponential moving averages. Optimization again uses \texttt{Adam} ($3\times10^{-4}$, clipping at $1.0$) over 150 episodes of 300 steps each (episodes are longer than in stage one to accommodate the scheduled perturbation windows). Because the goal is to track the evolution of $g$, training and evaluation are conducted simultaneously in this stage, and $g$ is carried across episode boundaries within each run.

\subsection{Perturbation Design}
\par Perturbations are implemented as transient observation-level distortions scheduled within each episode. During a perturbation window (15 steps), an inversion pattern is applied to the local patch $x_t$, making left-side cells appear less noisy and right-side cells more noisy, thereby inverting the apparent predictability gradient. The distortion is additive with fixed scale $0.12$; window onsets are scheduled within each episode with Gaussian timing jitter (s.d.\ 5 steps) around evenly spaced anchors, and are capped so that no window overlaps an episode boundary.

\par Crucially, the underlying environmental structure (grid scaffold, mean-value field, noise map and its transition dynamics) remains fixed throughout, so perturbations act only at the level of sensory appearance. The number of perturbation windows per episode is denoted $n_P$. Because the scaffold remains fixed, repeated perturbations allow testing whether transient distortions leave persistent residue in the agent's latent dynamics after conditions are restored.

\subsection{Evaluation Conditions} 
\par During the second stage of training, three primary evaluation methods are used, each designed to test a different aspect of history-dependent perceptual reorganization.

\paragraph{Probe-based representation assay.} To test whether the perspective latent $g$ directly reorganizes perceptual encoding, a fixed probe set of 25 observations (5 sampled positions per spatial zone, deterministic seed) is constructed from the environment under no-perturbation setting. For each completed mixed-history run ($n_P = 0 \!\to\! 4 \!\to\! 0$), late-block perspective vectors are extracted by averaging the latent state $g$ over the final 10 episodes of Block~0 and Block~2, yielding $g_0$ (pre-perturbation) and $g_2$ (post-perturbation recovery). A null condition $g=0$ is included as a reference.

\par Each probe observation $x$ is passed through the frozen encoder base MLP to obtain the raw perceptual code defined in Eq.~\ref{eq1}, which is identical across conditions. The FiLM salience gate then applies a $g$-specific modulation (Eqs.~\ref{eq2}--\ref{eq3}). Because the base MLP is frozen, any difference in the resulting $z_t$ is attributable solely to the FiLM modulation induced by $g$. In the main probe comparison, representations encoded under $g_2$ are compared against the null condition $g=0$ using PCA projection and per-dimension signed differences $z_t(g_2)-z_t(g=0)$ (Fig.~\ref{fig:probe}(a), (b)). Separately, the FiLM coefficients $\gamma$ are compared between $g_0$ and $g_2$ to test whether identical environmental conditions are encoded differently after intervening perturbation history (Fig.~\ref{fig:probe}(c)).

\paragraph{Plasticity ablation.} To distinguish the contribution of self-modulating plasticity from generic recurrent dynamics, the \textit{Adaptive} update law ($\alpha_t \in [0.03, 0.30]$) is compared against two fixed-rate controls under matched perturbation conditions ($n_P\!=\!4$): a \textit{Rigid} regime (constant $\alpha = 0.05$) and an \textit{Open} regime (constant $\alpha = 0.80$). The adaptive condition without perturbation ($n_P\!=\!0$) serves as an additional baseline (Fig.~\ref{fig:ablation}). This comparison is designed to test whether the qualitative dynamics of selective reopening and restabilization require adaptive self-modulation or arise from any sufficiently slow (or fast) update mechanism.

\paragraph{Mixed-history perturbation comparison.} This experiment compares plasticity dynamics across three block schedules, each consisting of three 50-episode blocks (150 episodes total). In the \textit{Mixed} perturbation condition ($n_P = 0 \!\to\! 4 \!\to\! 0$), perturbed episodes are present only in the middle block, allowing a direct comparison between the first and third blocks, which present identical environmental conditions but differ in intervening experience. To control for natural stabilization over training, two reference conditions are drawn from constant-perturbation runs: a \textit{Baseline} condition ($n_P = 0 \!\to\! 0 \!\to\! 0$; no perturbation throughout all 150 episodes) and a \textit{Persistent} perturbation condition ($n_P = 4 \!\to\! 4 \!\to\! 4$; full perturbation throughout all 150 episodes). Mean $\alpha$ is computed over the same episode windows in all three conditions, so that any difference in the magnitude of plasticity decrease reflects the effect of perturbation history rather than training duration.

\paragraph{Statistical analysis.} Descriptive summaries use the hierarchical median-of-medians scheme defined at the start of the Results section. Inference on the residue comparison uses a linear mixed-effects model on per-run block-wise plasticity change, with condition as a fixed effect and first-stage seed as a random intercept, so that the non-independence of runs sharing a first-stage backbone is respected. The cross-run reproducibility of salience-gating changes is assessed per dimension with two-sided binomial tests on the direction of change across the 30 mixed-history runs, Bonferroni-corrected for the 16 gating dimensions; a dimension is reported as reorganized only if it additionally exceeds a magnitude floor of $|\tilde{\Delta\gamma}| \geq 0.02$, preventing near-zero dimensions from registering as directionally consistent.

\par The geometry of the probe encodings is characterized by three quantities computed per run over the 25 probe encodings and then aggregated. Total encoding variance, $\mathrm{tr}\,\mathrm{cov}(z)$, measures overall gain. Because gain alone cannot distinguish reorganization from uniform amplification, two scale-invariant measures are reported alongside it: the participation ratio $(\sum_i \lambda_i)^2 / \sum_i \lambda_i^2$, which gives the effective number of dimensions over which encoding variance is distributed, and the angle between the leading principal axes of the encodings under the compared perspective states. Both are unchanged by any uniform rescaling of the encoding, so a change in either isolates differential reweighting across channels.

\section{Results and Analysis}

\par Conditions are evaluated hierarchically using two seed levels: 5 first-stage training seeds (0--4), each paired with 6 second-stage training seeds (0--5), which consists of 30 runs per condition. Summary statistics are computed by first taking the median across second-stage seeds within each first-stage seed, then summarizing across first-stage seeds using the median and interquartile range (IQR).

\subsection{Perspective Reorganizes Perception}
\begin{figure*}[t]
\centering
\includegraphics[width=1.0\textwidth]{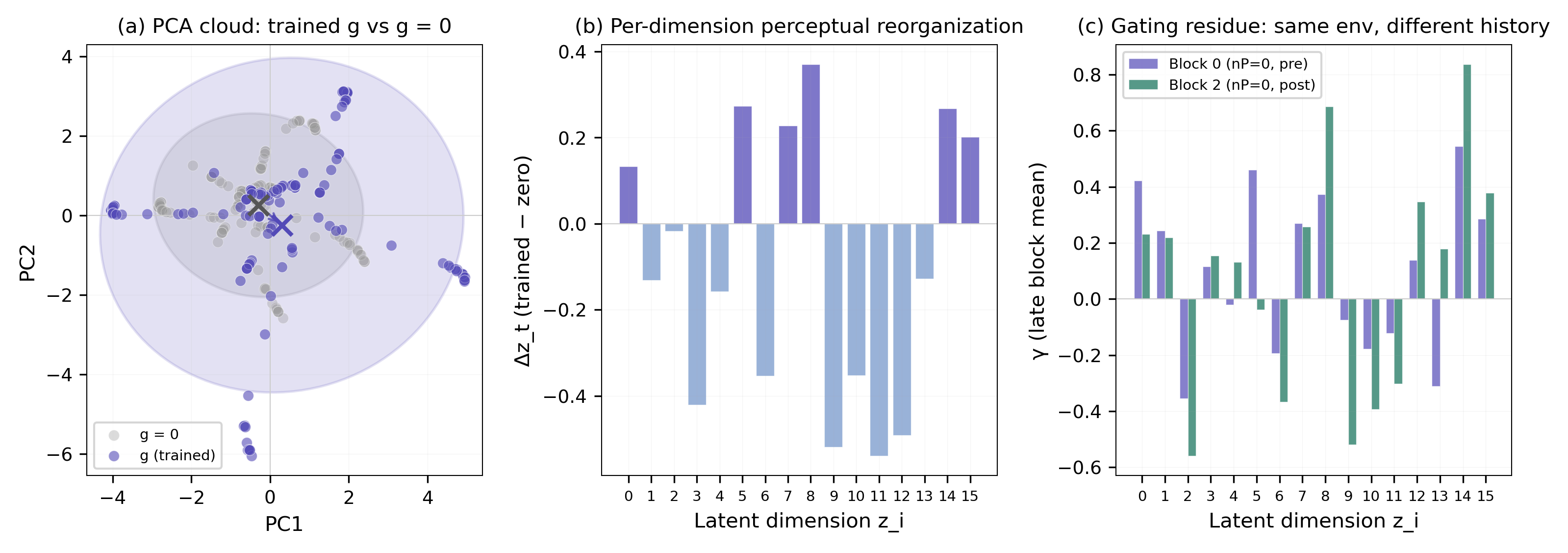}
\caption{\textbf{Perspective reorganizes perception of the same input.} The comparison perspective $g$ was obtained from a single illustrative mixed perturbation run ($n_P = 0\!\to\!4\!\to\!0$); cross-run statistics are reported in the text. \textbf{(a)} PCA projection of probe encodings under the late post-perturbation perspective state $g_2$ and the null condition $g\!=\!0$, shown with covariance ellipses. Cluster centroids are marked with crosses. Encoding-geometry statistics (variance ratio, participation ratio, principal-axis rotation) are computed across all 30 runs and reported in the text. \textbf{(b)} Signed per-dimension difference $z_t(g_2) - z_t(g\!=\!0)$ averaged across probe observations. \textbf{(c)} Per-dimension FiLM salience-gating coefficients $\gamma$ in Block 0 (pre-perturbation, $n_P\!=\!0$) and Block 2 (post-perturbation, $n_P\!=\!0$) of the mixed-history run.}
\label{fig:probe}
\end{figure*}

\par If the perspective latent merely stored past representations, changes in $g$ would not affect perception $z$ itself; the same observation would always produce the same encoding. Perspectival perception makes a stronger claim: the encoding of an observation depends partly on the perspective state under which it is encountered. The probe assay tests this claim directly. Because the encoder base MLP is frozen, any difference in $z_t$ across conditions must result from $g$-dependent FiLM modulation. The three panels of Fig.~\ref{fig:probe} examine this effect from complementary angles.

\par Fig.~\ref{fig:probe}(a) compares PCA projections of $z_t$ vectors obtained from fixed probe observations under two conditions: a history-conditioned perspective extracted from a mixed-history run, and a null perspective. Two features of the encoding geometry are informative, and both were quantified across all 30 mixed-history runs rather than read off the illustrated run. First, the encoding is amplified: the ratio of total encoding variance under $g_2$ to that under the null condition is $1.67$ (IQR $1.57$--$2.00$), exceeding one in 30 of 30 runs. 

\par However, a uniform gain applied to every channel would produce the same amplification without reorganizing anything. The second feature discriminates between these possibilities. The leading principal axis of the encoding rotates by a median of $25.5^\circ$ (IQR $23.1^\circ$--$28.6^\circ$), exceeding $5^\circ$ in 30 of 30 runs with a minimum of $14.7^\circ$. Because the FiLM transformation is diagonal and the probe covariance is invariant to the additive term $\beta$, the encoding covariance takes the form $D\,\mathrm{cov}(z_{\mathrm{raw}})\,D$ with $D = \mathrm{diag}(1+\gamma)$: a uniform gain leaves the principal axes exactly unchanged, so any rotation reflects differential scaling across perceptual channels.

\par The probe set elicits a predominantly one-dimensional response from the frozen encoder---roughly 89\% of the encoding variance lies along a single axis (participation ratio $1.12$)---so this rotation applies to the dominant encoding direction itself, with approximately 44\% of the post-perturbation axis lying outside the original one. The encoding also becomes marginally more concentrated under $g_2$ (participation ratio $1.12 \rightarrow 1.07$; decreasing in 29 of 30 runs, binomial $p = 5.8\times10^{-8}$), though this shift is small relative to the rotation. Taken together, the perspective latent redistributes variance across perceptual channels rather than merely rescaling or shifting the encoding.

\par Fig.~\ref{fig:probe}(b) shows the signed per-dimension difference $z_t(g_2) - z_t(g=0)$, averaged across probe observations. The perspective latent selectively amplifies certain dimensions (notably $z_5$, $z_7$, $z_8$) while suppressing others (such as $z_3$, $z_9$, $z_{11}$). This pattern emerges from the learned interaction between the spatial gradient structure, the FiLM parameterization, and the accumulated perspective dynamics. It provides direct evidence that changes at the level of $g$ propagate into the perceptual layer $z_t$, ultimately altering how the same observation $x_t$ is encoded.

\par Roughly put, panels (a) and (b) can be understood as a form of $\mathrm{do}(g)$ intervention. Panel (c) goes a step further: Fig.~\ref{fig:probe}(c) compares the per-dimension salience gating pattern $\gamma$ between Block 0 and Block 2 of a mixed-history run. Although both blocks present identical environmental conditions, the late-block gating patterns differ substantially, indicating that the intervening perturbation experience has restructured how the salience gate modulates individual latent channels.

\par Across all 30 mixed-history runs, the direction of the Block~0 $\rightarrow$ Block~2 change in $\gamma$ was tallied per dimension. 5 of the 16 gating dimensions change direction consistently across runs after Bonferroni correction ($\alpha = 0.05$) while exceeding the magnitude floor: $z_9$ shifts in the same direction in 29 of 30 runs (corrected $p = 9.2\times10^{-7}$), $z_2$ and $z_6$ in 26 of 30 (corrected $p \approx 1.0\times10^{-3}$), and $z_{14}$ and $z_4$ in 25 of 30 (corrected $p \approx 5\times10^{-3}$), with median magnitudes $|\tilde{\Delta\gamma}|$ between $0.13$ and $0.40$. 

\par Thus, history-driven changes in salience gating were reproducible across runs. In several dimensions, the direction of change was consistent even though $g$ developed through each agent's own episode history. Other dimensions also changed substantially, but not in a consistent direction. This suggests that the architecture produces both a shared pattern of reorganization and additional run-specific variation. 

\par In sum, Fig.~\ref{fig:probe} establishes that $g$ intervenes on $z$: perception varies as a function of perspective in this agent architecture. The perspective latent does not merely drift as a passive summary of past experience; it \textbf{actively reorganizes how the same world is given to the agent}. How the latent comes to support such reorganization depends on its own update dynamics, which the next analysis dissects.

\subsection{Plasticity Shows a Distinct Update Regime}
\begin{figure}[t]
\centering
\includegraphics[width=1.0\columnwidth]{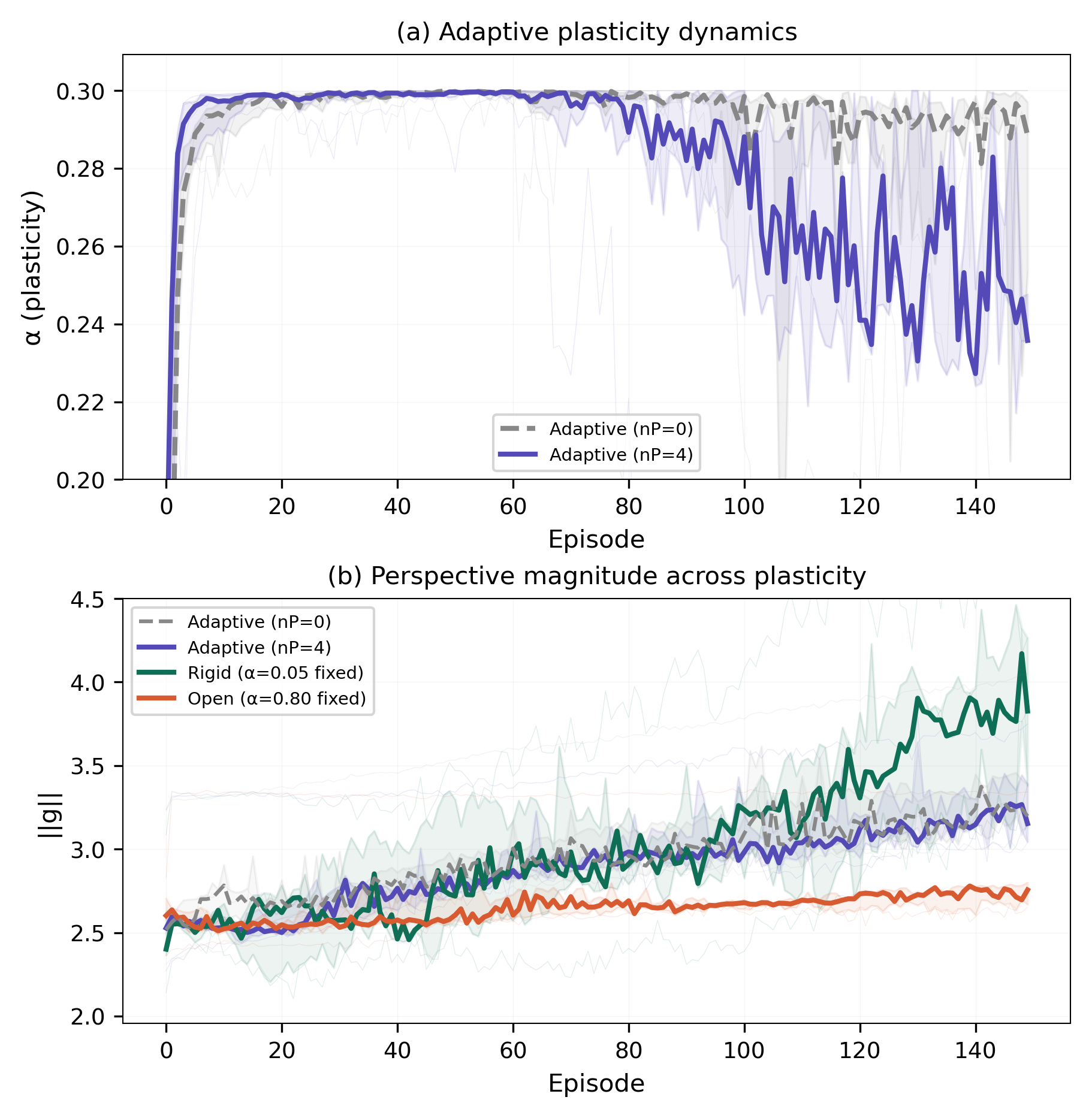} 
\caption{\textbf{Ablation analysis of self-modulating plasticity.} \textbf{(a)} Episode-wise adaptive plasticity $\alpha$ with perturbation ($n_P=4$) and without perturbation ($n_P=0$). In both conditions, $\alpha$ rises rapidly early in training, but only the perturbed condition later declines below baseline. \textbf{(b)} Perspective magnitude $\|g\|$ across four update regimes: Adaptive with perturbation, Adaptive baseline, Rigid ($\alpha=0.05$), and Open ($\alpha=0.80$). The rigid regime shows continued growth, the open regime remains relatively flat, and the adaptive regimes lie between these extremes, with slightly larger late-stage $\|g\|$ under perturbation.}
\label{fig:ablation}
\end{figure}

\par The update dynamics themselves discriminate between hypotheses. If perspectival organization were reducible to generic recurrence, a fixed update rate (slow or fast) should reproduce the same latent dynamics. If self-modulation matters, the regimes should differ qualitatively. Fig.~\ref{fig:ablation}(a) compares the trajectory of $\alpha$ under self-modulating plasticity with and without perturbation. In both conditions, plasticity is initially high during early learning. The divergence emerges later; under the perturbation-free baseline, $\alpha$ remains elevated, whereas under repeated perturbation it progressively declines with high variance. This suggests that adaptive plasticity is a history-sensitive variable that diminishes over time as structure consolidates under repeated exposure.

\par Fig.~\ref{fig:ablation}(b) compares $\|g\|$ across Adaptive, Rigid, and Open update laws. Under the Rigid regime ($\alpha\!=\!0.05$), the GRU candidate nudges $g$ at each step, but the low update rate preserves most of the prior state. The result is a persistent unidirectional drift: $\|g\|$ grows without bound, indicating a failure of self-regulation in which structure accumulates without stabilization. Under the Open regime ($\alpha\!=\!0.80$), roughly 80\% of $g$ is overwritten by the new GRU output at every step, leaving little trace of prior history. The latent magnitude remains comparatively flat, reflecting a reactive system that encodes the present moment rather than accumulating perspective over time. The Adaptive regime falls between these extremes: plasticity is initially high, allowing $\|g\|$ to grow as representational structure is acquired, and then decreases as the system stabilizes, producing a growth-then-plateau trajectory. Notably, the adaptive conditions with and without perturbation ($n_P\!=\!4$ and $n_P\!=\!0$) converge to similar late-training $\|g\|$ values. This suggests that while perturbation may alter \textit{how} the perspective latent is organized, it does not substantially change the total amount of structure that $g$ accumulates.

\par Since $g$ is zero-initialized, $\|g\|$ can be understood as a scalar summary of how far the latent has moved from its origin, i.e. a proxy for the total accumulated perspective structure, though it may be agnostic to the direction of that accumulation. What matters in Fig.~\ref{fig:ablation}(b), therefore, is not the absolute value of $\|g\|$ but the \textit{shape} of its trajectory: unbounded drift under rigid updating, stationarity under open updating, and growth-then-stabilization under adaptive self-modulation. These three qualitatively distinct regimes confirm that \textbf{self-modulating plasticity defines a structurally different mode of latent dynamics}. The divergence of $\alpha$ under perturbation in panel (a) already suggests that plasticity carries history; the following analysis tests this directly under matched conditions.

\subsection{Perturbation History and Adaptive Plasticity}
\begin{figure}[t]
\centering
\includegraphics[width=1.0\columnwidth]{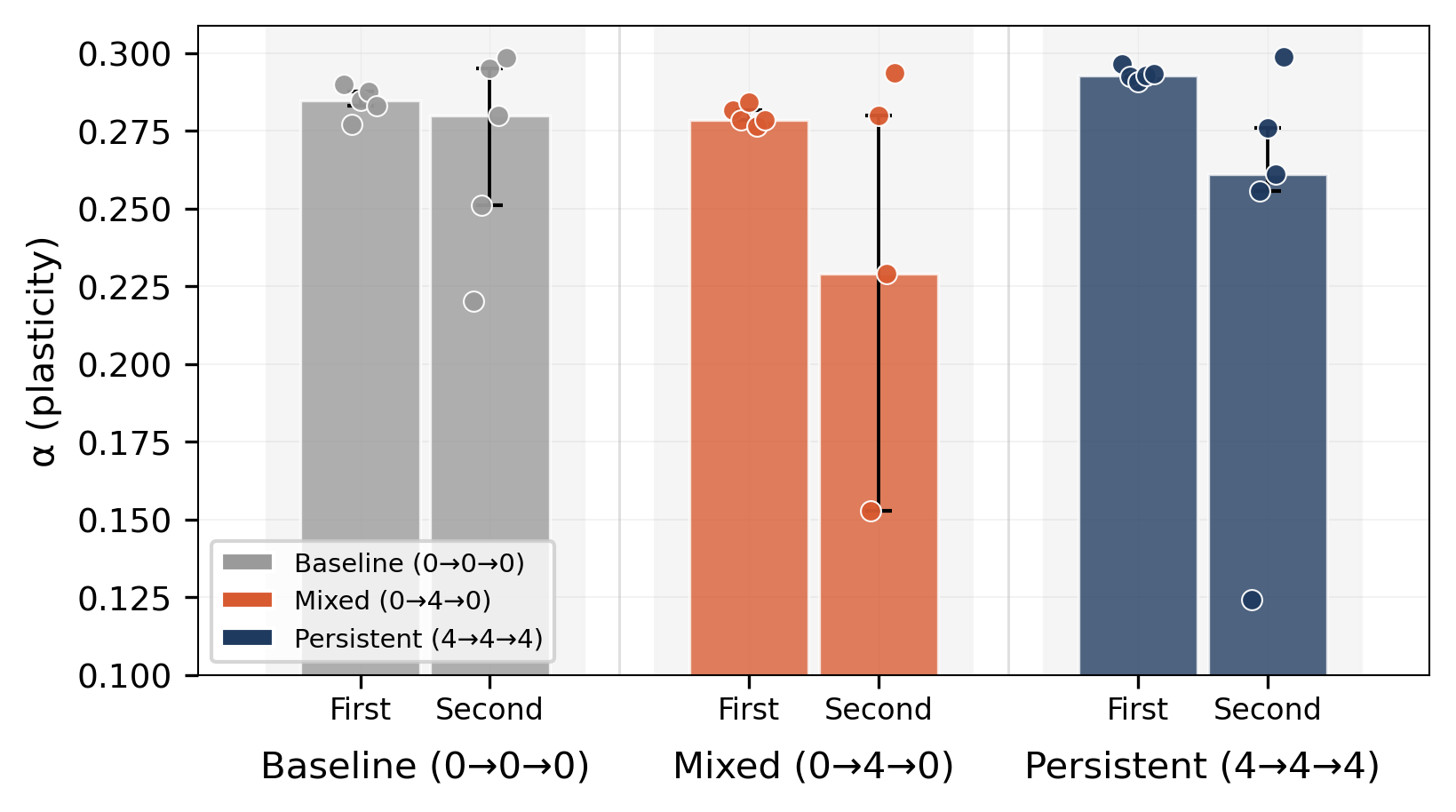} 
\caption{\textbf{Plasticity under matched three-block histories.} Mean adaptive plasticity $\alpha$ is compared between the first and final blocks across three schedules matched in total duration: Baseline ($n_P = 0\!\to\!0\!\to\!0$), Mixed perturbation ($n_P = 0\!\to\!4\!\to\!0$), and Persistent perturbation ($n_P = 4\!\to\!4\!\to\!4$). Bars show hierarchical medians, points show individual seed run values, and error bars indicate IQR. The block-wise decline is smallest under Baseline ($0.285$ to $0.280$), largest under Mixed ($0.279$ to $0.229$), and comparable under Persistent ($0.293$ to $0.261$); see text for inferential statistics.}
\label{fig:residue}
\end{figure}

\par If the perspective latent were a passive summary of recent input, plasticity should depend only on current conditions: two no-perturbation blocks separated by intervening perturbation should look no different from two separated by quiet training. If history is retained as organization, the intervening experience should leave a trace after conditions are restored. Fig.~\ref{fig:residue} tests this by comparing the first and third blocks across the three schedules. The theoretically diagnostic comparison is Mixed versus Baseline: only in these conditions are the compared blocks nominally identical ($n_P = 0$ in both), so that any difference is attributable to intervening history rather than to concurrent perturbation.

\par Descriptively, the ordering matches the history-residue prediction. The hierarchical median decline in $\alpha$ between the matched blocks is $0.003$ under Baseline, $0.044$ under Mixed, and $0.037$ under Persistent. A linear mixed-effects model with first-stage seed as a random intercept estimates the Mixed-Baseline difference at $+0.028$ (95\% CI $[-0.003, +0.058]$, $p = 0.075$), with 4 of the 5 first-stage seed clusters showing the predicted direction. The trend is thus consistent, though not statistically resolved; with 5 first-stage seeds, the analysis is underpowered for an effect of this size, hence I report it as directional evidence rather than a confirmed difference. The Mixed and Persistent declines are statistically indistinguishable ($p = 0.93$); since the Persistent condition's third block contains ongoing perturbation, its decline confounds current conditions with history, which is why the Mixed-Baseline contrast carries the inferential weight.

\par The distribution of late-block plasticity is more informative than its mean. Rather than shifting uniformly, runs tend toward one of two attractors: strong consolidation ($\alpha < 0.20$ in the final block) or persistence near the adaptive ceiling ($\alpha > 0.295$, i.e., at the upper bound $\alpha_{\max} = 0.30$). Intervening perturbation shifts the balance toward consolidation: 11 of 30 Mixed runs consolidate against 4 of 30 Baseline runs (odds ratio $3.8$, Fisher's exact $p = 0.072$), while ceiling-pinned runs are correspondingly rarer under Mixed (7/30 vs. 12/30). This bimodality also explains the high across-seed variance of the mean: for runs pinned at $\alpha_{\max}$, self-modulation is saturated and history cannot register in the plasticity rate at all, capping the measurable effect. Hence, perturbation history appears to act as a switch toward consolidation rather than as a graded shift.

\subsection{Behavioral Stability Under Perceptual Reorganization}
\par Why does the perceptual reorganization documented above not translate into changed gross behavior? Across mixed-history runs, spatial behavior in Block~2 closely matches Block~0: the median total-variation distance between zone-occupancy distributions is $0.030$ (IQR $0.010$--$0.100$; $0$ = identical), and the median shift in mean column position is $+0.10$ cells. The residual movement that does occur is itself orderly: runs already occupying the low-noise region (19 of 30) show essentially no change, while the remainder drift further toward the low-noise side (median $+1.8$ columns) - movement that continues the orientation established in the first training stage rather than departing from it. 

\par Two factors jointly account for this stability, and both are informative. First, it is partly by construction: the policy and state head are frozen in the second stage, so behavioral reorganization is structurally damped - which is a deliberate choice that allows perceptual reorganization to be measured without being confounded by concurrent policy adaptation. Second, it is not entirely trivial: the frozen policy still acts on the gated percept $z_t$ and on $g_t$ through $s_t$, so a sufficiently large perceptual reorganization \textit{could} have propagated into behavior, and did not. Within this architecture, accumulated perspective reorganizes how the world is encoded while leaving spatial habits intact. Whether stronger behavioral consequences emerge when the backbone is unfrozen is an open question for future work.

\subsection{Summary of the Results}
\par Three conclusions follow from the analyses. First, the perspective latent $g$ reorganizes the perceptual latent $z$: identical observations are encoded differently depending on prior experience, and the reorganization of salience gating replicates across independent runs (Fig.~\ref{fig:probe}). Second, only adaptive self-modulation exhibits the growth-then-stabilization dynamic, distinguishing it qualitatively from rigid and open update regimes (Fig.~\ref{fig:ablation}). Third, intervening perturbation history is followed by reduced plasticity between nominally identical blocks, which is a directionally consistent trend that manifests as an increased tendency toward consolidation (Fig.~\ref{fig:residue}). Throughout the simulation, the agent's behavior remains stable, localizing the reorganization at the perceptual level.

\section{Discussion}

\subsection{Two Architectural Pillars: Perspectival Perception and Self-Regulating Maturity}
\par The results converge on two coupled architectural contributions that define the model's approach to perspectival organization. The first is \textit{perspectival perception}: the same nominal observation can be encoded differently depending on the observer's accumulated stance. This is implemented through the FiLM layer, which transforms $z_{\mathrm{raw}}$ as a function of $g$, such that the perceptual representation $z_t$ is structurally conditioned by experiential history (Fig.~\ref{fig:probe}(a)--(b)). In phenomenological terms, what changes is not the \textit{stimulus} itself but the \textit{mode of its disclosure}. The world does not change; the way it is given to the agent does.

\par The second is \textit{self-regulating maturity}: openness to revision is not fixed externally but determined in part by the system's own state. Here, the term ``maturity'' is used in a restricted sense borrowed from constructive-developmental psychology \cite{kegan1982}, denoting the capacity to balance openness to revision against the stability of accumulated organization - while neither collapsing into rigidity nor into perpetual reactivity. In the model, this capacity corresponds directly to the measured $\alpha$ dynamic: initially high plasticity, progressive consolidation, and selective reopening under perturbation (Fig.~\ref{fig:ablation}).

\par Most importantly, \textbf{these two pillars are inseparable.} Perspectival perception without self-regulating maturity would be unstable, as each new input would reorganize perception without constraint. By the same token, self-regulating maturity without perspectival perception would be inert, as the system could modulate its own plasticity without any downstream consequence for how the world is taken up. The feedback loop between $g$ and $z_t$---perspective shapes perception, and gated perception in turn drives the perspective update---gives the architecture its characteristic dynamics. Neither component, taken alone, would let history bear on how the world is subsequently encountered.

\subsection{Maturity, Perceptual Reorganization, and Their Interdependence}
\par This two-pillar structure resonates with a theme in adult developmental psychology and contemplative science, where maturity and perceptual reorganization are understood as interconnected aspects of development. In Kegan's constructive-developmental framework, each stage of adult development is marked by a reorganization of what the subject takes as given versus what it can reflect upon \cite{kegan1982}. Contemplative science reports analogous dynamics in advanced meditation practice. For example, the ``stages of insight'' involve progressive reorganization of perception, including shifts in precision weighting, attentional control, and most importantly, self-world boundaries \cite{yang2025stages}. Murray et al. describe this as the ``deautomation'' of cognitive structures: the loosening of habitual perceptual organization to permit reorganization at a higher level of integration \cite{murray2026}.

\par Across these accounts, maturity can be understood as the capacity to reorganize one's perceptual and interpretive stance in response to accumulated evidence without collapsing into rigidity or instability. The present architecture provides a minimal computational analog of this capacity. The FiLM layer implements perceptual reorganization, while AlphaNet implements self-regulation of openness. Because the two are coupled through feedback, neither develops in isolation; perception is reorganized only insofar as the perspective latent has matured, and the perspective latent matures only insofar as reorganized perception feeds back into its update.

\subsection{Beyond Belief State Tracking and Generic Recurrence}
\par A natural objection would be that any system with recurrent internal state exhibits history-dependent processing: in the classic contrast between reactive and dynamical agents \cite{beer2003}, a reactive agent is a function of its current sensory input, whereas a dynamical agent possesses a state space through which the history of its interactions shapes present responses. Since the perspective latent $g$ is maintained by a recurrent module, one may ask whether ``perspectival perception'' amounts to anything more than recurrent state dependence under a phenomenological label.

\par However, the claim of this study is more specific: the ablation results locate the present analysis within a spectrum-like distinction. The \textit{Open} regime occupies the reactive pole, while the \textit{Rigid} regime lies precisely at a generic slowly-updating recurrent system pole. Neither extreme reproduces the observed dynamics: the first accumulates no perspective, the second accumulates it without ever stabilizing (Fig.~\ref{fig:ablation}). Only when the update rate is itself a bounded, state-dependent variable does the history-dependent profile appear by combining initial openness, progressive consolidation, and selective reopening under perturbation. 

\par A sharper version of the objection would note that gated recurrent architectures already contain state-dependent update modulation: a GRU's update gate interpolates between prior state and candidate state as a learned function of input and hidden state. Three differences matter. First, in a standard gated cell, update modulation is entangled with content computation - that is, per-dimension gates trained end-to-end inside the same pathway that computes the new state. Here, plasticity is factored out as a separate scalar channel: a dedicated network receives the current percept, action trace, prior perspective, and an explicit error-feedback vector, and returns a single bounded rate $\alpha_t$ applied outside the recurrent candidate computation (Eqs.~\ref{eq4}--\ref{eq5}). This separation follows work on neuromodulated plasticity, where a distinct modulatory signal regulates how much learning occurs rather than what is learned \cite{miconi2019,najarro2020}. Second, because $\alpha_t$ is a scalar, openness to revision becomes a measurable state variable in its own right - which is exactly what makes the residue analyses of this study possible: the equivalent quantity in a standard gated cell is distributed across dimensions and not straightforwardly interpretable as a degree of openness. Third, and most importantly, the modulated state feeds back into perception. In a generic recurrent agent, hidden state conditions downstream computation---what is done with the input---while the encoding of the input itself remains a fixed feedforward stage. Here, $g$ reaches back into the encoder through the FiLM gate, so that history changes the representation of the input. It is this very closed loop through perception that the term \textit{perspectival} is meant to mark.

\par The comparison with belief states sharpens the same point from a normative angle. In a Partially Observable Markov Decision Process (POMDP), the belief state is a sufficient statistic for the current hidden state, updated to track incoming evidence as accurately as possible \cite{pomdp1998,spaan2012}. For an ideal filter, dependence on the path by which the current posterior was reached is a defect: two agents holding the same posterior should behave identically regardless of how they arrived at it. Residue, a persistent difference in internal organization after nominal conditions are restored, is therefore a \textit{bug} for a belief state and a defining \textit{feature} of the perspective latent. The ablation makes this concrete: the fast-updating regime, which most closely approximates responsive evidence-tracking, is exactly the regime in which no perspectival structure accumulates. A belief state supports current-state estimation, where the perspective latent supports history-dependent perceptual orientation. The two should be understood as complementary functional roles, not as competing implementations of the same role.

\par This reading connects naturally to a more general Active Inference framework, where perception is shaped by precision weighting---the gain assigned to prediction errors---and where regime-level context is carried by slowly evolving hidden states \cite{kirchhoff2017,linson2018}. Salience gating can be viewed as a learned, stance-conditioned analog of precision modulation over perceptual channels, and self-modulating plasticity as precision control applied to belief updating itself. A formal treatment of $g$ as a hierarchical precision state within an active inference generative model is a natural next step.

\par A further structural point separates $g$ from the hidden states of standard recurrent policy architectures. In recurrent reinforcement learning agents, the hidden state is shaped by return-maximizing gradients and is therefore, functionally, an instrument of the policy. However, in this study, actor gradients are blocked from $g$ by its architectural construction: the policy operates on a stop-gradient copy of the state, and $g$ is shaped exclusively by predictive coherence and its own temporal regularization. Whatever organization $g$ acquires is thus not an artifact of action optimization, which licenses interpreting it as a perceptual-organizational variable. 

\par Nevertheless, the scope of this argument should be stated honestly. These are architectural and dynamical distinctions, demonstrated in a minimal setting under matched conditions. A sufficiently large, end-to-end-trained recurrent network could in principle emulate the observed dynamics, with the relevant structure entangled across its weights. The contribution of the present architecture is to isolate the ingredients---that is, a slow latent decoupled from action optimization, a scalar self-modulated plasticity, and a feedback path into perceptual encoding---in a form where each is independently measurable and ablatable. On this reading, the question is not whether generic recurrence \textit{could} realize perspectival organization, but what minimal and interpretable structure suffices to produce and study it.

\subsection{Toward Perspectival Artificial Agents} 
\par Current language-based AI systems can display elements of cognitive empathy and emotionally supportive language, yet evaluations also note prompt sensitivity, repetitive empathic phrasing, and inconsistency across contexts \cite{sorin2024}. Such systems may reproduce the discursive surface of empathy without instantiating the more primitive, history-dependent form of attunement that phenomenological and affective accounts associate with genuine perspective \cite{ajeesh2025,colombetti2014}. The present architecture suggests one possible direction for what such attunement would minimally require. If perspectival organization requires at minimum (1) a slow latent that resists rapid revision, (2) a feedback loop through which that latent reorganizes perceptual encoding, and (3) a self-modulating mechanism that regulates openness to change, then these properties are unlikely to emerge from optimization on text corpora alone. They require an agent situated in a world, encountering perceptual perturbation over time, with internal dynamics structured both to permit and resist reorganization. This echoes arguments in embodied cognitive architectures that affective and regulatory dynamics must be constitutive rather than post-hoc features of the architecture \cite{ziemke2009}.

\par Importantly, this should not be taken as a criticism of language-based models. Rather, the proposal is that perspectival organization can provide a complementary layer that could coexist with many forms of policy-adaptation machinery, including those used in LLM-based systems. Perspective operates along a different functional axis and need not compete with behavioral adaptation. The two-stage training protocol already illustrates this separation. In the first stage, behavioral policy adaptation is established as a distinct process of entropy-minimizing adjustment to environmental structure. In the second stage, perspective is introduced not as a replacement, but as a higher-order-like organization that shapes how such adaptation is guided and sustained over time. On this view, perspectival organization and policy adaptation are not opposing alternatives, but potentially coexisting components of a richer artificial agent architecture.

\subsection{Limitations and Future Directions}
\par The main inferential limitation is the seed budget. With 5 first-stage seeds as the effective clustering unit, the residue comparison is underpowered: the effect is directionally consistent in four of 5 clusters, but its confidence interval includes zero. The measurable effect is additionally capped by the plasticity bound itself (i.e. runs pinned at $\alpha_{\max}$ cannot express history in their plasticity rate at all), so the reported residue should be regarded as a lower bound on the underlying tendency. Expanding the number of first-stage seeds is the most direct extension.

\par Several aspects of the design space were deliberately left unexplored. The dependence of history effects on perturbation intensity, window length, and scheduling was not systematically characterized; all main analyses use $n_P = 4$ with 15-step windows. The behavioral backbone was frozen in the second stage, which damps behavioral consequences by construction; whether perceptual reorganization propagates into behavior when the backbone is unfrozen is an open empirical question. A salience-gating-off control ($\gamma = \beta = 0$ fixed) would further isolate the FiLM pathway's contribution to the system-level dynamics. 

\par The probe set also elicits a predominantly one-dimensional response from the frozen encoder, so the perceptual reorganization measured here concerns the rotation and rescaling of a single dominant encoding axis; a probe set constructed to span a higher-dimensional response manifold would test whether the same mechanism produces richer restructuring. The environment is deliberately minimal (an 8-dimensional local observation over a fixed scaffold), and first-stage backbones are themselves heterogeneous - one of the 5 seeds stabilized near the high-noise edge rather than the low-noise region; all analyses retain all seeds, and the hierarchical and mixed-effects statistics are robust to this heterogeneity, but generalization to richer environments remains to be shown.

\par Lastly, the interpretive scope should be delimited. As stated at the outset, the measured quantities are computational analogs of perspectival organization, and not a claim that the agent instantiates perspective in the strictly phenomenological sense.

\section{Conclusion}

\par This paper introduces a minimal architecture in which a slow perspective latent $g$ shapes perception $z$ through salience gating, and regulates its own plasticity. The results show that $g$ reorganizes the encoding of identical observations, and that only adaptive self-modulation yields growth followed by stabilization; intervening perturbation history is further followed by reduced plasticity, a directionally consistent trend under the present seed budget. Together, these findings identify a minimal, ablatable mechanism by which experiential history becomes an organizing condition for subsequent experience.

\section*{Data Availability}

\par The code and data supporting the findings of this study are publicly available at \url{https://github.com/hjpae/cearlab-phase2}.

\bibliographystyle{mslapa}
\bibliography{references}

\end{document}